\documentclass[runningheads]{llncs}
\usepackage{graphicx}
\usepackage{amsmath,amssymb} 
\usepackage{color}
\usepackage[width=122mm,left=12mm,paperwidth=146mm,height=193mm,top=12mm,paperheight=217mm]{geometry}

\usepackage{floatrow}
\newfloatcommand{capbtabbox}{table}[][\FBwidth]

\begin{document}
\pagestyle{headings}
\mainmatter

\title{Deep Learning of Local RGB-D Patches for 3D Object Detection and 6D Pose Estimation} 

\titlerunning{Deep Learning of Local RGB-D Patches for Detection and Pose Estimation}

\authorrunning{Kehl et al. }

\author{Wadim Kehl $^\dag$, Fausto Milletari $^\dag$, Federico Tombari $^{\dag \S}$, \\ Slobodan Ilic $^{\dag *}$, Nassir Navab $^\dag$}


\institute{Technical University of Munich $^\dag$ \hspace{0.3cm}  University of Bologna $^\S$  \hspace{0.3cm}   Siemens AG, Munich $^*$ }

\maketitle

\begin{abstract}
We present a 3D object detection method that uses regressed descriptors of locally-sampled RGB-D patches for 6D vote casting. For regression, we employ a convolutional auto-encoder that has been trained on a large collection of random local patches. During testing, scene patch descriptors are matched against a database of synthetic model view patches and cast 6D object votes which are subsequently filtered to refined hypotheses. We evaluate on three datasets to show that our method generalizes well to previously unseen input data, delivers robust detection results that compete with and surpass the state-of-the-art while being scalable in the number of objects.  
\end{abstract}

\section{Introduction}

Object detection and pose estimation are of primary importance for tasks such as robotic manipulation, scene understanding and augmented reality, and have been the focus of intense research in recent years. The availability of low-cost RGB-D sensors enabled the development 
of novel methods that can infer scale and pose of the object more accurately even in presence of occlusions and clutter. 

\begin{figure}[b]
	\centering
	\includegraphics[width=4cm]{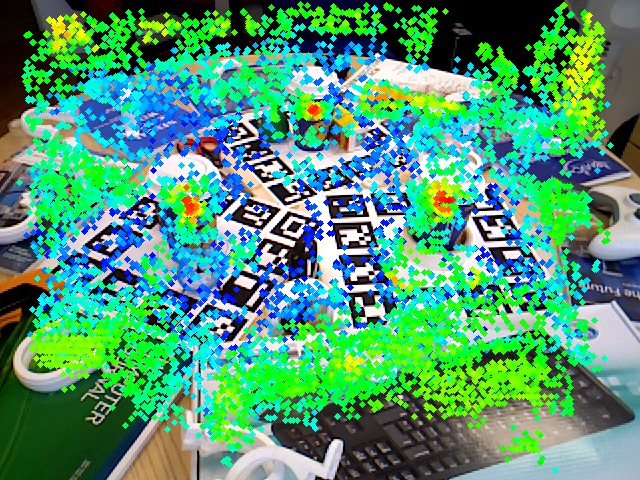}
	\includegraphics[width=4cm]{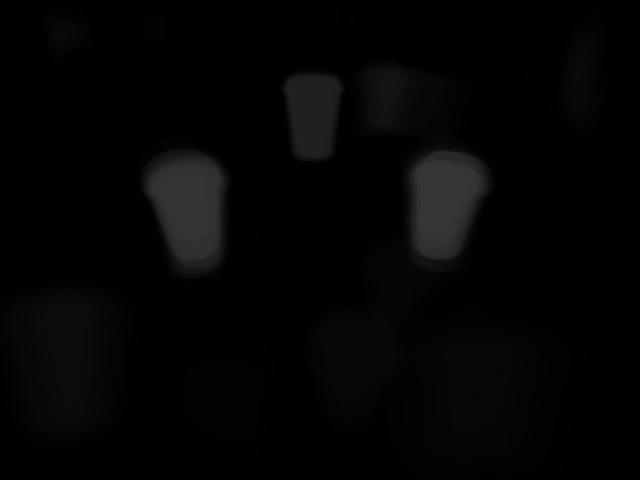}
	\includegraphics[width=4cm]{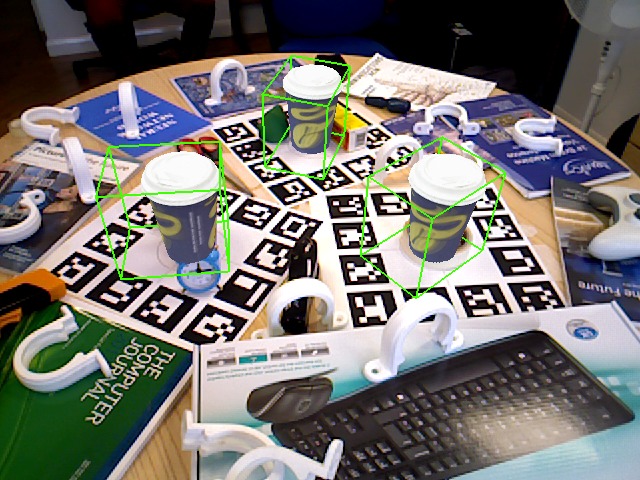}
	\caption{
		Results of our voting-based approach that uses auto-encoder descriptors of local RGB-D patches for 6-DoF pose hypotheses generation. (Left) Cast votes from each patch indicating object centroids, colored with their confidence. (Middle) Segmentation map obtained after vote filtering. (Right) Final detections after pose refinement.}
	\label{fig:teaser}
\end{figure}

Methods such as Hinterstoisser et al. and related \cite{Hinterstoisser2012a,Rios-Cabrera2013,Kehl2015} detect objects in the scene by employing templates generated from synthetic views and matching them efficiently against the scene. While these holistic methods are implemented to be very fast at a low FP-rate, their recall drops quickly in presence of occlusion or substantial noise. Differently, descriptor-based approaches \cite{Mian2009,Hao2013,Aldoma2013} rely on robust schemes for correspondence grouping and hypothesis verification to withstand occlusion and clutter, but are computationally intensive.
Other methods like Brachmann et al. \cite{Brachmann2014} and Tejani et al. \cite{Tejani2014} follow a local approach where small RGB-D patches vote for object pose hypotheses in a 6D space. Although such methods are not taking global context into account, they proved to be robust towards occlusion and the presence of noise artifacts since they infer the object pose using only its parts. Their implementations are based on classical Random Forests where the chosen features to represent the data can strongly influence the amount of votes that need to be cast to accomplish the task and, consequently, the required computational effort.

Recently, convolutional neural networks (CNNs) have shown to outperform state-of-the-art approaches in many computer vision tasks by leveraging the CNNs' abilities of automatically learning features from raw data. CNNs are capable of representing images in an abstract, hierarchical fashion and once a suitable network architecture is defined and the corresponding model is trained, CNNs can cope with a large variety of object appearances and classes.

Recent methods performing 3D object detection and pose estimation successfully demonstrated the use of CNNs on data acquired through RGB-D sensors such as depth or normals. For example, \cite{Gupta2014,Gupta2015} make use of features produced by a network to perform classification of region proposals via SVMs. A noteworthy work is Wohlhart et al. \cite{Wohlhart2015}, that demonstrates the applicability of CNNs for descriptor learning of RGB-D views. This work uses a holistic approach and delivers impressive results in terms of object retrieval and pose estimation, although can not be directly applied to object detection in clutter since a precise object localization would be needed. Nonetheless, it does hint towards replacing hand-crafted features with learned ones for this task.

Our work is inspired by~\cite{Wohlhart2015} and we demonstrate that neural networks coupled with a local voting-based approach can be used to perform reliable 3D object detection and pose estimation under clutter and occlusion. To this end, we deeply learn descriptive features from local RGB-D patches and use them afterwards to create hypotheses in the 6D pose space, similar to~\cite{Brachmann2014,Tejani2014}. 

\begin{figure}[h]
	\centering
	\includegraphics[width=12cm]{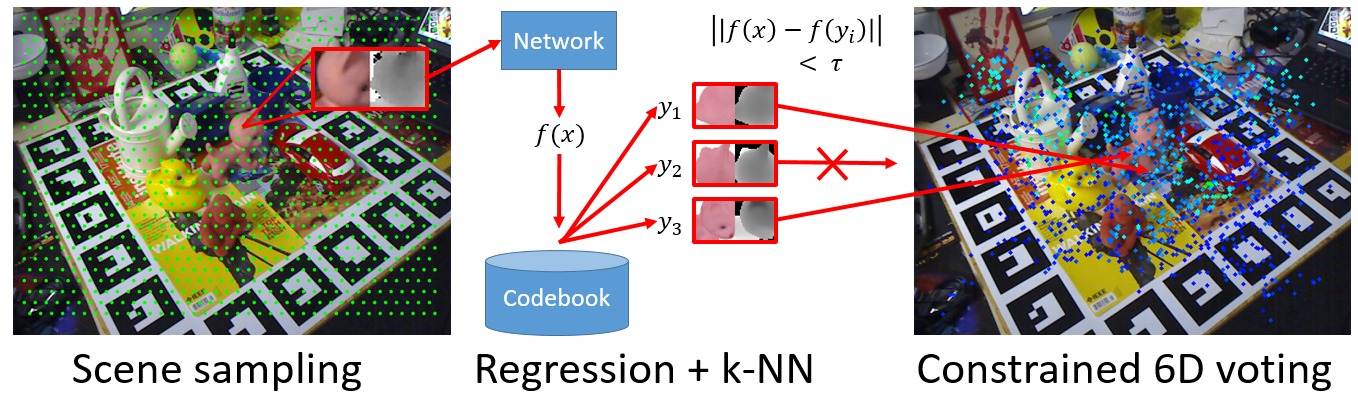}
	\caption{Illustration of the voting. We densely sample the scene to extract scale-invariant RGB-D patches. These are fed into a network to regress features for a subsequent k-NN search in a codebook of pre-computed synthetic local object patches. The retrieved neighbors then cast 6D votes if their feature distance is smaller than a threshold $\tau$.}
	\label{fig:pipeline}
\end{figure}

In practice, we train a convolutional autoencoder (CAE) \cite{Masci2011} from scratch using random patches from RGB-D images with the goal of descriptor regression. With this network we create codebooks from synthetic patches sampled from object views where each codebook entry holds a local 6D pose vote. In the detection phase we sample patches in the input image on a regular grid, compute their descriptors and match them against codebooks with an approximate k-NN search. Matching returns a number of candidate votes which are cast only if their matching score surpasses a threshold (see a schematic in Figure \ref{fig:pipeline}).

We will show that our method allows for training on real data, efficient matching between synthetic and real patches and that it generalizes well to unseen data with an extremely high recall. Furthermore, we avoid explicit background learning and scale well with the number of objects in the database.

\section{Related work}

There has recently been an intense research activity in the field of 3D object detection, with many methods proposed in literature traditionally subdivided into feature-based and template-based. As for the first class, earlier approaches relied on features \cite{Lowe2004,Bay2006} directly detected on the RGB image and then back-projected to 3D~\cite{Lowe2001,Pauwels2013}. With the introduction of 3D descriptors \cite{Rusu2009,Tombari2010}, approaches replaced image features with features directly computed on the 3D point cloud \cite{Mian2009}, and introduced robust schemes for filtering wrong 3D correspondences and for hypothesis verification \cite{Hao2013,Aldoma2013,Buch2014}. They can handle occlusion and are scalable in the number of models, thanks to the use of approximate nearest neighbor schemes for feature matching~\cite{Muja2014} yielding sub-linear complexity. Nevertheless, they are limited when matching surfaces of poor informative shape and tend to report non real-time run-times. 

On the other hand, template-based approaches are often very robust to clutter but scale linearly with the number of models. LineMOD~\cite{Hinterstoisser2012} performed  robust 3D object detection by matching templates extracted from rendered views of 3D models and embedding quantized image contours and normal orientations. Successively, \cite{Rios-Cabrera2013} optimized the matching via a cascaded classification scheme, achieving a run-time increase by a factor of 10. Improvements in efficiency are also achieved by the two-stage cascaded detection method in~\cite{Cai2013} and by the hashing matching approach tailored to LineMOD templates proposed in \cite{Kehl2015}. Other recent approaches \cite{Malisiewicz2011,Gu2010,Aubry2014} build discriminative models based on such representations using SVM or boosting applied to training data. 

Recently, another category of methods has emerged based on \emph{learning} RGB-D representations, which are successively classified or matched at test time. \cite{Brachmann2014,Tejani2014} use random forest-based voting schemes on local patches to detect and estimate 3D poses. While the former regresses object coordinates and conducts a subsequent energy-based pose estimation, the latter bases its voting on a scale-invariant LineMOD-inspired patch representation and returns location and pose simultaneously. Recently, CNNs have also been employed \cite{Wohlhart2015,Gupta2014,Gupta2015} to learn RGB-D features. 
The main limitations of this category of methods is that, being based on discriminative classifiers, they usually require to learn the background as a negative class, thus making their performance dataset-specific. Instead, we train neural networks in an unsupervised fashion and use them as a plug-in replacement for methods based on local features.

\section{Methodology}

In this section, we first give a description of how we sample local RGB-D patches of the given target objects and the scene while ensuring scale-invariance and suitability as a neural network input. Secondly, we describe the employed neural networks in more detail. Finally, we present our voting and filtering approach which efficiently detects objects in real scenes using a trained network and a codebook of regressed descriptors from synthetic patches.

\subsection{Local Patch Representation}

Our method follows an established paradigm for voting via local information. Given an object appearance, the idea is to separate it into its local parts and let them vote independently \cite{Pepik2012,Gall2011,Brachmann2014,Tejani2014}. While most approaches rely on hand-crafted features for describing these local patches, we tackle the issue by regressing them with a neural network.

\begin{figure}
	\centering
	\includegraphics[width=4.5cm]{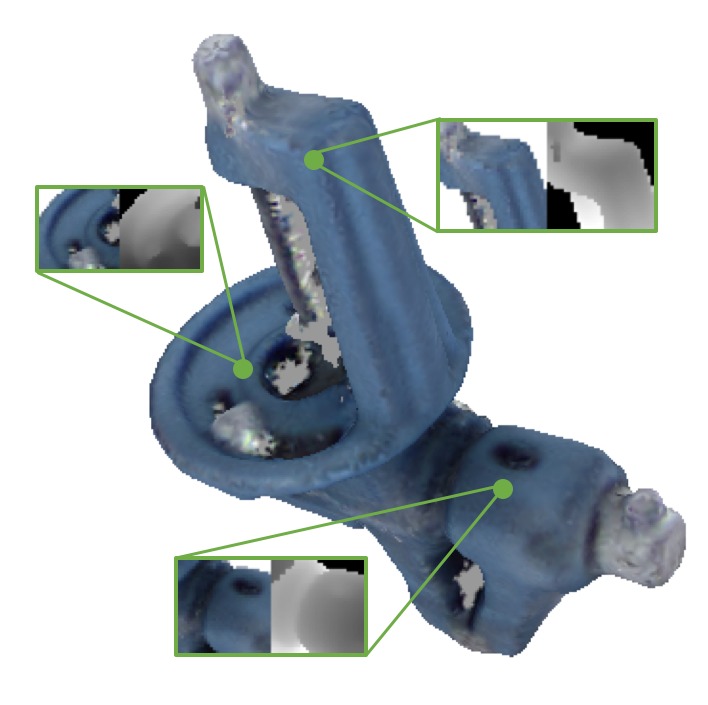}
	\hspace{1cm}
	\includegraphics[width=6.5cm]{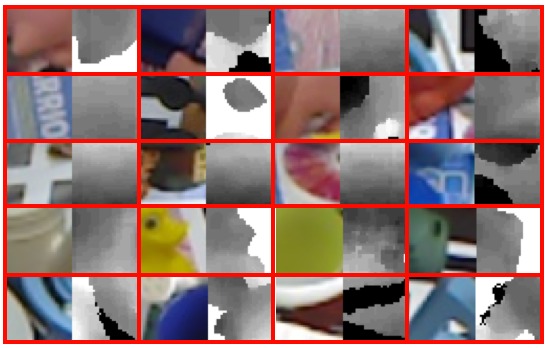}
	\caption{Left: For each synthetic view, we sample scale-invariant RGB-D patches $\mathbf{y}_i$  of a fixed metric size on a dense grid. Their associated regressed features $f(\mathbf{y}_i)$ and local votes $v(\mathbf{y}_i)$ are stored into a codebook. Right: Examples from the approx. 1.5 million random patches taken from the LineMOD dataset for autoencoder training. }
	\label{fig:patches}
\end{figure}

To represent an object locally, we render it from many viewpoints equidistantly sampled on an icosahedron (similar to \cite{Hinterstoisser2012}), and densely extract a set of scale-independent RGB-D patches from each view. To sample invariantly to scale, we take depth $z$ at the patch center point and compute the patch pixel size such that the patch corresponds to a fixed metric size $m$ (here: 5 cm) via
\begin{equation}
patch_{size} =  \frac{m}{z} \cdot f
\end{equation}
with $f$ being the focal length of the rendering camera. After patch extraction, we de-mean the depth values with $z$ and clamp them to $\pm m$  to confine the patch locally not only along x and y, but also along z. Finally, we normalize color and depth to $[-1,1]$ and resize the patches to $32 \times 32$. See Figure \ref{fig:patches} for an exemplary synthetic view together with sampled local patches.

An important advantage of using local patches as in the proposed framework is that it avoids the problematic aspect of background modeling. Indeed, for what concerns discriminative approaches based on learning a background and a foreground class, a generic background appearance can hardly be modeled, and recent approaches based on discriminative classifiers such as CNNs deploy scene data for training, thus becoming extremely dataset-specific and necessitating refinement strategies such as hard negative mining. Also, both \cite{Brachmann2014} and \cite{Wohlhart2015} model a supporting plane to achieve improved results, with the latter even introducing real images intertwined with synthetic renderings into the training to force the CNN to abstract from real background. Our method instead does not need to model the background at all.

\subsection{Network Training}
Since we want the network to produce discriminative features for the provided input RGB-D patches, we need to bootstrap suitable filters and weights for the intermediate layers of the network. Instead of relying on pre-trained, publicly available networks, we decided to train from scratch due to multiple reasons:

\begin{enumerate}
	\item Not many works have incorporated depth as an additional channel in networks and most remark that special care has to be taken to cope with, among others, sensor noise and depth 'holes' which we can control with our data.
	\item We are one of the first to focus on local RGB-D patches of small-scale objects. There are no pre-trained networks that have been so far learned on such data, and it is unclear how well other networks that were learned on RGB-D data can generalize to our specific problem at hand.
	\item To robustly train deep architectures, a high amount of training samples is needed. By using patches from real scenes, we can easily create a huge training dataset which is specialized to our task, thus enhancing the discriminative power of our network.
\end{enumerate}

Note that other works usually train a CNN on a classification problem and then use a 'beheaded' version of the network for other tasks (e.g. \cite{Girshick2014}). Here, we cannot simply convert our problem into a feasible classification task because of the sheer amount of training samples that range in the millions. Although we could assign each sample to the object class it belongs to, this would bias the feature training and hence, counter the learning of a generalized patch feature representation, independent of object affiliations. It is important to point out that also \cite{Wohlhart2015} aimed for feature learning, but with a different goal. Indeed, they enforce distance similarity of feature space and object pose space, while we instead strive for a compact representation of our local input patches, independent of the objects' poses.

We teach the network regression on a large variety of input data by randomly sampling local patches from the LineMOD dataset \cite{Hinterstoisser2012}, amounting to around 1.5 million total samples. Furthermore, these samples were augmented such that each image got randomly flipped and its color channels permutated. Our network aims to learn a mapping from the high-dimensional patch space to a much lower feature space of dimensionality $F$, and we employ a traditional autoencoder (AE) and a convolutional autoencoder (CAE) to accomplish this task. 

\begin{figure}[t]
	\centering
	\includegraphics[width=8cm]{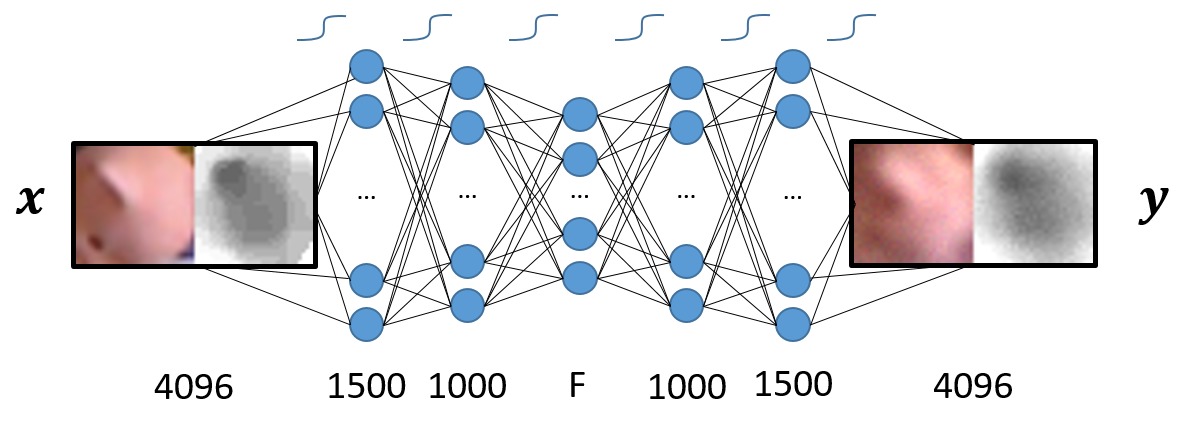}
	\vspace{0.05cm}
	\hrule
	\vspace{0.1cm}
	\includegraphics[width=10cm]{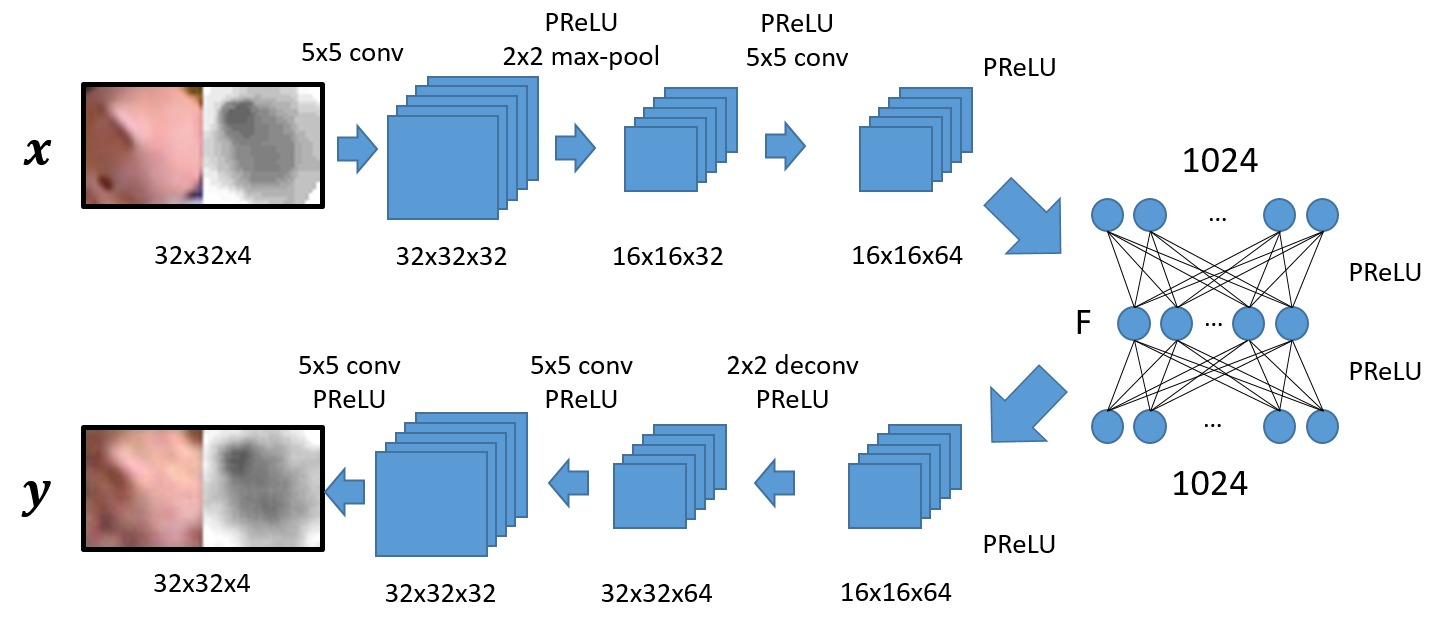}
	\caption{Depiction of the employed AE (top) and CAE (bottom) architectures. For both, we have the compressing feature layer with dimensionality $F$.}
	\label{fig:networks} 
\end{figure}

Autoencoders minimize a reconstruction error $||x-y||$ between an input patch $x$ and a regressed output patch $y$ while the inner-most compression layer condenses the data into $F$ values. We use these $F$ values as our descriptor since they represent the most informative compact encoding of the input patch. Our architectures can be seen in Figure \ref{fig:networks}. For the AE we use two encoding and decoding layers which are all connected with tanh activations. For the CAE we employ multiple layers of $5\times 5$ convolutions and PReLUs (Parametrized Rectified Linear Unit) before a single fully-connected encoding/decoding layer, and use a deconvolution with learned $2 \times 2$ kernels for upscaling before proceeding back again with $5 \times 5$ convolutions and PReLUs. Note that we conduct one max-pool operation after the first convolutions to introduce a small shift-invariance.

\subsection{Constrained Voting}

\begin{figure}[t]
	\centering
	\includegraphics[width=4cm]{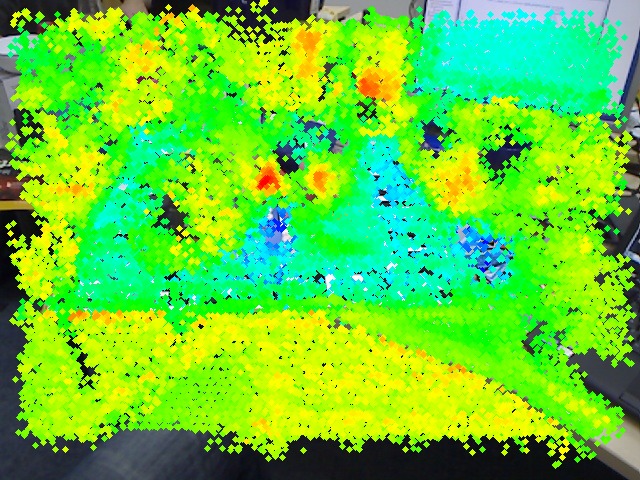}
	\includegraphics[width=4cm]{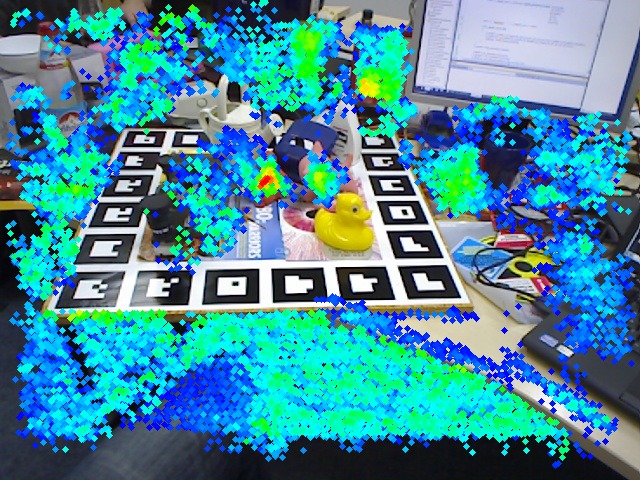}
	\includegraphics[width=4cm]{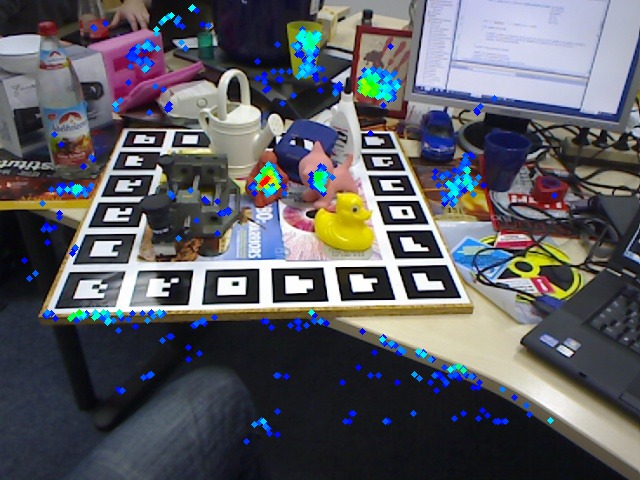}
	\caption{Casting the constrained votes for $k=10$ with a varying distance threshold (left to right):  $\tau=15$, $\tau=7$, $\tau=5$. The projected vote centroids $v_i$ are colored according to their scaled weight $ w(v_i) / \tau$. It can be seen that many votes accumulate confidently around the true object centroid for differently chosen thresholds.}
	\label{fig:votes}
\end{figure}

A problem that is often encountered in regression tasks is the unpredictability of output values in the case of noisy or unseen, ill-conditioned input data. This is especially true for CNNs as a deep cascade of non-linear functions composed of many parameters. In our case, this can be caused by e.g., unseen object parts, general background appearance or sensor noise in color or depth. If we were to simply regress the translational and rotational parts, we would be prone to this input sensitivity. Furthermore, this approach would always cast votes at each image sampling position, increasing the complexity of sifting through the voting space afterwards.  
Instead, we render an object from many views and store local patches $\mathbf{y}$ of this synthetic data in a database, as seen in Figure \ref{fig:patches}. For each $\mathbf{y}$, we compute its feature $f(\mathbf{y}) \in \mathbb{R}^F$ and store it together with a local vote $(t_x,t_y,t_z,\alpha,\beta,\gamma)$ describing the patch 3D center point offset to the object centroid and the rotation with respect to the local object coordinate frame. This serves as an object-dependent codebook. 

During testing, we take each sampled 3D scene point $\mathbf{s} = (s_x,s_y,s_z)$ with associated patch $\mathbf{x}$, compute its deep-regressed feature $f(\mathbf{x})$ and retrieve $k$ (approximate) nearest neighbors $\mathbf{y}_1,...,\mathbf{y}_k$. Each neighbor casts then a global vote $v(\mathbf{s},\mathbf{y}) = (t_x+s_x,t_y+s_y,t_z+t_y,\alpha,\beta,\gamma)$ with an associated weight $w(v)= e^{-||f(\mathbf{x})-f(\mathbf{y})||}$ based on the feature distance.

Notably, this approach is flexible enough to provide three main practical advantages. 
First, we can vary $k$ in order to steer the amount of possible vote candidates per sampling position. Together with a joint codebook for all objects, we can retrieve the nearest neighbors with sub-linear complexity, enabling scalability.
Secondly, we can define a threshold $\tau$ on the nearest neighbor distance, so that retrieved neighbors will only vote if they hold a certain confidence. This reduces the amount of votes cast over scene parts that do not resemble any of the codebook patches. Furthermore, if noise sensitivity perturbs our regressed feature, it is more likely to be hindered from vote casting. 
Lastly and of significance, it is assured that each vote is numerically correct because it is unaffected by noise in the input data, given that the feature matching was reliable. See Figure \ref{fig:votes} for a visualization of the constrained voting.

\subsubsection{Vote filtering}
\begin{figure}
	\centering
	\includegraphics[width=4cm]{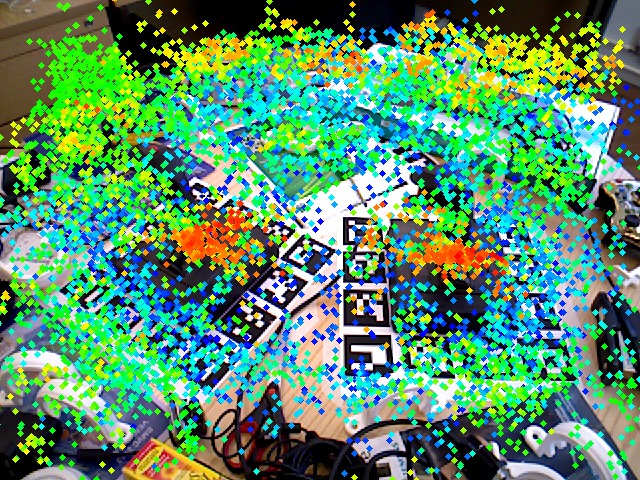}
	\includegraphics[width=4cm]{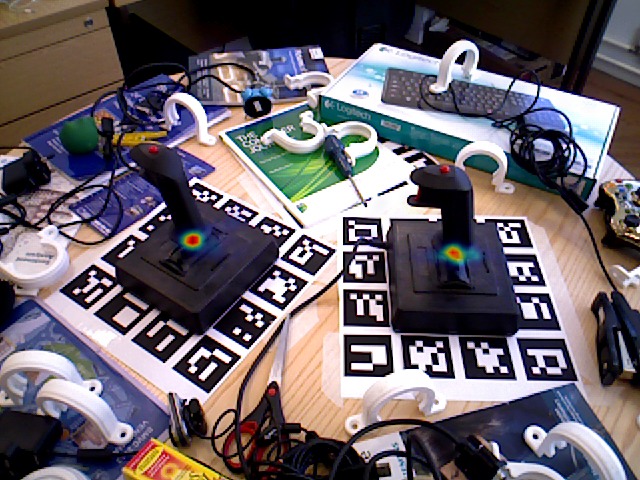}
	\includegraphics[width=4cm]{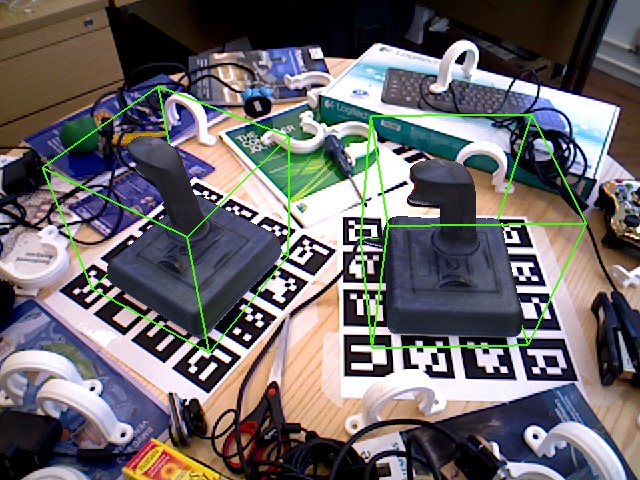}
	\caption{Starting with thousands of votes (left) we run our filtering to retrieve intermediate local maxima (middle) that are further verified and accepted (right).}
	\label{fig:filtering}
\end{figure}

Casting votes can lead to a very crowded vote space that requires refinement in order to keep detection computationally feasible. We thus employ a three-stage filtering: in the first stage we subdivide the image plane into a 2D grid (here: cell size of $5 \times 5$ pixels) and throw each vote into the cell the projected centroid points to. We suppress all cells that hold less than $k$ votes and extract local maxima after bilinear filtering over the accumulated weights of the cells. Each local mode collects the votes from its direct cell neighbors and performs mean shift with a flat kernel, first in translational and then in quaternion space (here: kernel sizes 2.5 cm and 7 degrees). This filtering is computationally very efficient and removes most spurious votes with non-agreeing centroids, while retaining plausible hypotheses, as can be seen in Figure \ref{fig:filtering}. Furthermore, the retrieved neighbors of each hypotheses' constituting votes hold synthetic foreground information that can be quickly accumulated to create meaningful segmentation maps (see Figure \ref{fig:teaser} for an example on another sequence).

\section{Evaluation}

\subsection{Reconstruction quality}
To evaluate the performance of the networks, we trained AEs and CAEs with feature layer dimensions $F \in \{  32,64,128,256 \}$. We implemented our networks with Caffe \cite{Jia2014} and trained each with an NVIDIA Titan X with a batch size of 500. The learning rate was fixed to $10 ^{-5}$ and we ran 100,000 iterations for each network. The only exception was the 256-dim AE, which we trained for 200,000 iterations for convergence due to its higher number of parameters.

\begin{figure}[t]
	\centering
	\includegraphics[width=12cm]{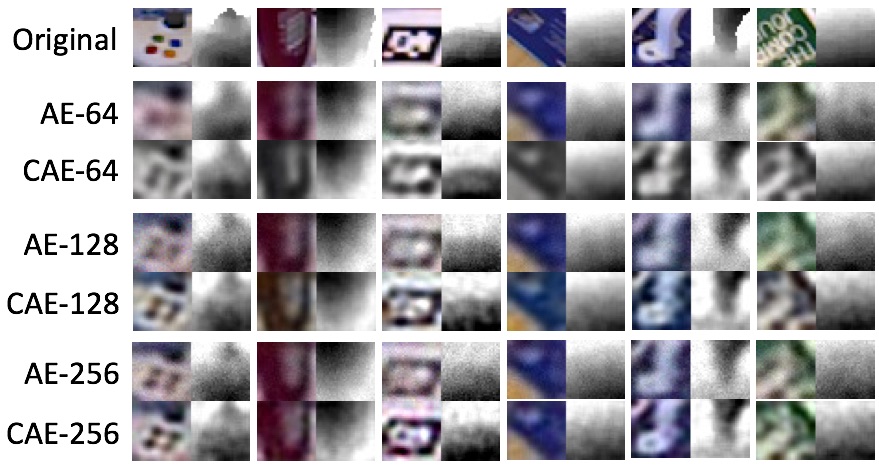}
	
	\caption{RGB-D patch reconstruction comparison between our AE and CAE for a given feature dimensionality $F$. Clearly, the AE and CAE focus on different qualities and both networks increase the reconstruction fidelity with a wider compression layer.}
	\label{fig:recons_img}
\end{figure}

For a visual impression of the results, we present the reconstruction quality side-by-side of AEs and CAEs on six random RGB-D patches in Figure \ref{fig:recons_img}. Note that these patches are test samples from another dataset and thus have not been part of the training, i.e. the networks are reconstructing previously unseen data. 

It is apparent that the CAEs put more emphasis on different image properties than their AE pendants. The AE reconstructions focus more on color and are more afflicted by noise since weights of neighboring pixels are trained in an uncorrelated fashion in this architecture. The CAE patches instead recover the spatial structure better at the cost of color fidelity. This can be especially seen for the 64-dimensional CAE where the remaining $1.56\% = (64/4096)$ of the input information forced the network to resort to grayscale in order to preserve image structure. It can be objectively stated that the convolutional reconstructions for 128 dimensions are usually closer to their input in visual terms. Subsequently, at dimensionality 256 the CAE results are consistently of higher fidelity both in terms of structure and color/texture.

\subsection{Multi-instance dataset from Tejani et al.}
We evaluated our approach on the dataset of Tejani et al. \cite{Tejani2014}. Upon inspection, the dataset showed problems with the ground truth annotation, which has been confirmed by the authors via personal communication. We thus re-annotated the dataset by ICP and visual inspection of each single frame. The authors then supplied us with their recomputed scores of the method in \cite{Tejani2014} on such corrected version of the dataset, which they are also going to publish on their website. To evaluate against the authors' method (LC-HF), we follow their protocol and extract the $N=5$ strongest modes in the voting space and subsequently verify each via ICP and depth/normal checks to suppress false positives.

We used this dataset first to evaluate how different networks and feature dimensions influence the final detection result. To this end, we fixed $k=3$ and conducted a run with the trained CAEs, AEs and also compared to PCA\footnote{Due to computational constraints we took only 1 million patches for PCA training. } as means for dimensionality reduction. Since different dimensions and methods lead to different feature distances we set $\tau = \infty$ for this experiment, i.e. votes are unconstrained. Note that we already show here generalization since the networks were trained on patches from another dataset. As can be seen in Table \ref{table:self}, PCA provides a good baseline performance that surpasses even the CAE at 32 dimensions, although this mainly stems from a high precision since vote centroids rarely agreed. In turn, both networks supplied similar behavior and we reached a peak at 128 with our CAE, which we fixed for further evaluation. We also found $\tau = 10$ and a sampling step of 8 pixels to provide a good balance between accuracy and runtime. For a more in-depth self comparison, we kindly refer the reader to the supplementary material.

For this evaluation we also supply numbers from the original implementation of LineMOD \cite{Hinterstoisser2012a}. Since LineMOD is a matching-based approach, we evaluated such that each template having a similarity score larger than 0.8 is taken into the same verification described above. It is evident that LineMOD fares very well on most sequences since the amount of occlusion is low. It only showed problems where objects sometimes are partially outside the image plane (e.g. 'joystick','coffe'), have many occluders and thus a smaller recall ('milk')  or where the planar 'juice' object decreased the precision by occasional misdetections in the table. Not surprisingly, LineMOD outperforms the other two methods largely for the small 'camera' since it searches the entire specified scale space whereas LC-HF and our method both rely on local depth for scale inference. Although our local voting does detect instances in the table as well, there is rarely an agreeing centroid that survives the filtering stage and our method is by far more robust to larger occlusions and partial views. We are thus overtaking the other methods in 'coffe' and 'joystick'. The 'milk' object is difficult to handle with local methods since it is uniformly colored and symmetric, defying a reliable accumulation of vote centroids. Although the 'joystick' is mostly black, its geometry allows us to recover the pose very reliably. All in all, we outperform the state-of-the art in holistic matching slightly while clearly improving over the state-of-the-art in local-based detection by significant 9.6\% on this challenging dataset. Detailed numbers are given in Tables 2 and 3. Unfortunately, runtimes for LC-HF are not provided by the authors.

\begin{table}
	\begin{center}
		\begin{tabular}{@{}c|c|c|c|c@{}}
			$F$    & 32    & 64    & 128   & 256 \\ \hline
			PCA  & 0.33 & 0.43 & 0.46 & 0.47
		\end{tabular}
		\hspace{0.2cm}
		\begin{tabular}{@{}c|c|c|c|c@{}}
			$F$ & 32    & 64    & 128   & 256 \\ \hline
			AE& \textbf{0.43} & \textbf{0.63} & 0.65 & 0.66
		\end{tabular}
		\hspace{0.2cm}
		\begin{tabular}{@{}c|c|c|c|c@{}}
			$F$ & 32    & 64    & 128   & 256 \\ \hline
			CAE & 0.32 & 0.58 & \textbf{0.70} & \textbf{0.69}
		\end{tabular}
	\end{center}
	\caption{F1-scores on the Tejani dataset using PCA, AE and CAE for patch descriptor regression with a varying dimension $F$. We highlight the best method for a given $F$. Note that the number for CAE-128 deviates from Table 3 since here we set $\tau = \infty$.}
	\label{table:self}
\end{table}

\begin{figure}[t]
	\centering
	\includegraphics[width=4cm]{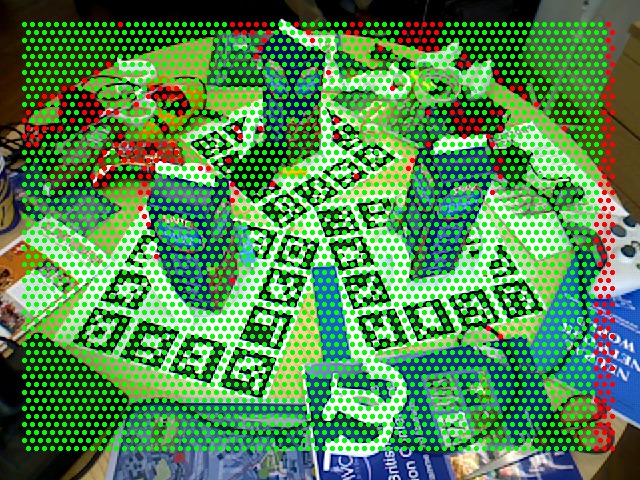}
	\includegraphics[width=4cm]{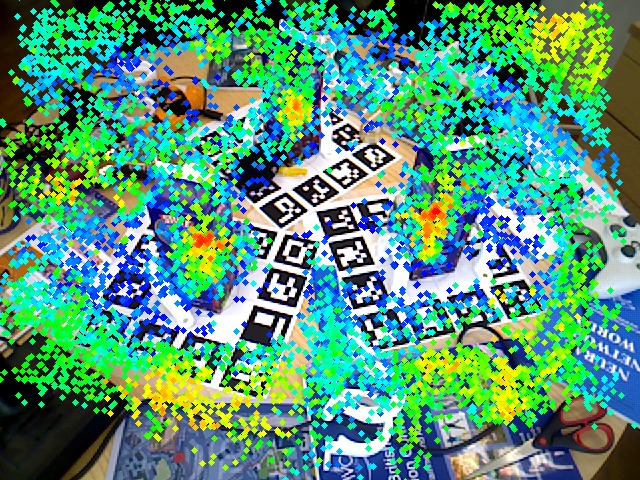}
	\includegraphics[width=4cm]{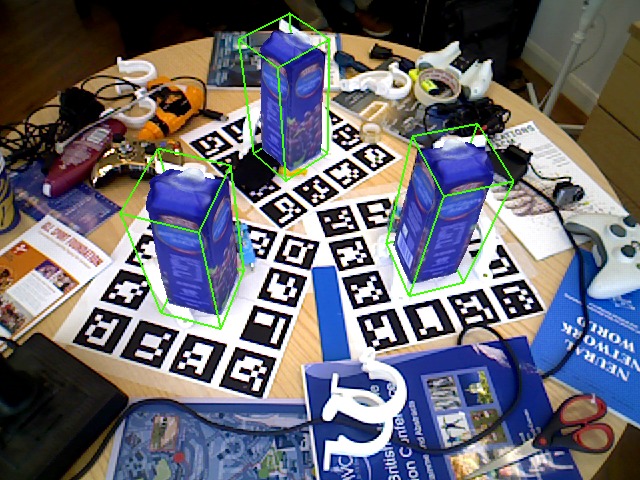}
	\includegraphics[width=4cm]{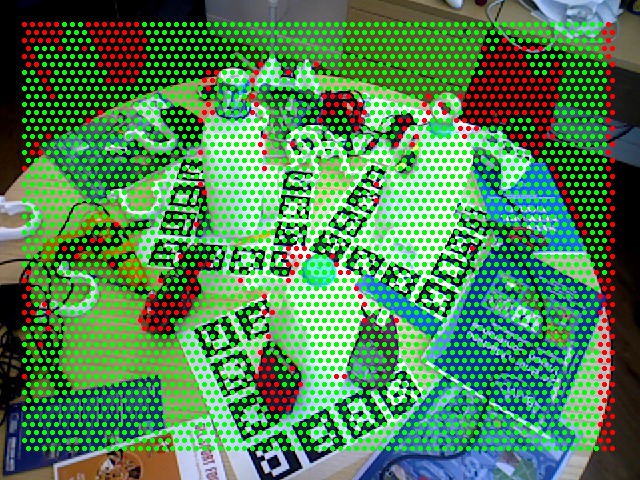}
	\includegraphics[width=4cm]{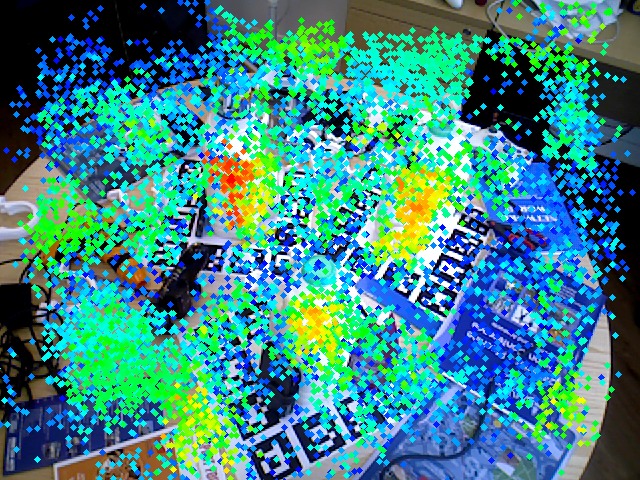}
	\includegraphics[width=4cm]{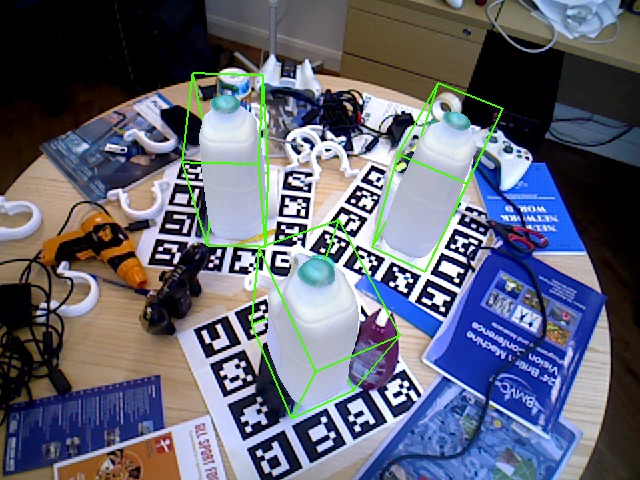}
	\caption{Scene sampling, vote maps and detection output for two objects on the Tejani dataset. Red sample points were skipped due to missing depth values.}
	\label{fig:tejani_img}
\end{figure}

\begin{figure}
	
	\begin{floatrow}
		
		\capbtabbox{
			
			\begin{tabular}{c|c}
				Stage			& Runtime (ms)  \\
				\hline	    
				Scene sampling 	& 0.03\\
				CNN regression 	& 477.3\\
				k-NN \& voting  & 61.4\\
				Vote filtering 	& 1.6 \\
				Verification	& 130.5\\
				\hline
				Total			& 670.8\\
			\end{tabular}
		}{\caption{Average runtime on \cite{Tejani2014}. Note that the feature regression is done on the GPU.}}
		
		\capbtabbox{
			\begin{tabular}{c|c|c|c}
				Sequence		& LineMOD & LC-HF	& Our approach \\
				\hline	    
				Camera (377) 	& \textbf{0.589} & 0.394 & 	0.383 \\
				Coffee (501)  	& 0.942 & 0.891 &	\textbf{0.972} \\
				Joystick (838)  & 0.846 & 0.549	& \textbf{0.892} \\
				Juice (556)		& 0.595	& \textbf{0.883}& 0.866 \\
				Milk (288)		& \textbf{0.558}& 0.397	& 0.463 \\
				Shampoo (604)	& \textbf{0.922}	& 0.792	& 0.910 \\
				\hline
				Total (3164)	&  0.740     & 0.651	& \textbf{0.747}\\
			\end{tabular}
		}{\caption{F1-scores for each sequence on the re-annotated version of \cite{Tejani2014}. Note that we show the updated LC-HF scores provided by the authors.}}			
		
	\end{floatrow}
\end{figure}

\subsection{LineMOD dataset}
We evaluated our method on the benchmark of \cite{Hinterstoisser2012} in two different ways. To compare to a whole set of related work that followed the original evaluation protocol, we remove the last stage of vote filtering and take the $N=100$ most confident votes for the final hypotheses to decide for the best hypothesis and use the factor $k_m=0.1$ in their proposed error measure. To evaluate against Tejani et al. we instead follow their protocol and extract the $N=5$ strongest modes in the voting space and choose $k_m=0.15$. Since the dataset provides one object ground truth per sequence, we use only the codebook that is associated to that object for retrieving the nearest neighbors. Two objects, namely 'cup' and 'bowl', are missing their meshed models which we manually created. For either protocol we eventually verify each hypothesis via a fast projective ICP followed by a depth and normal check. Results are given in Tables \ref{table:linemod_prec} and \ref{table:linemod_f1}. 

\begin{figure}
	\centering
	\includegraphics[width=3.7cm]{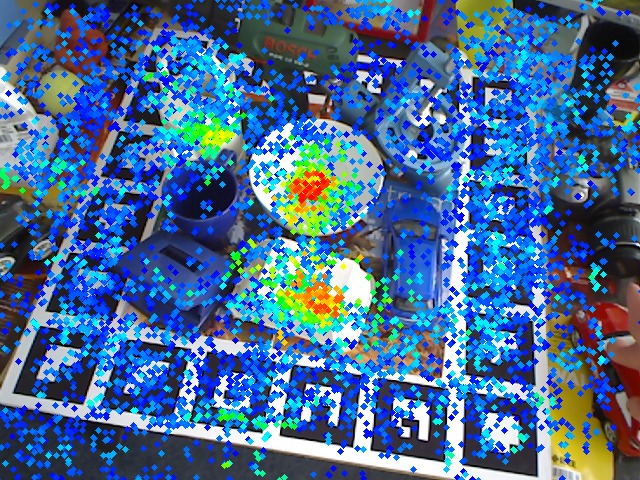}
	\includegraphics[width=3.7cm]{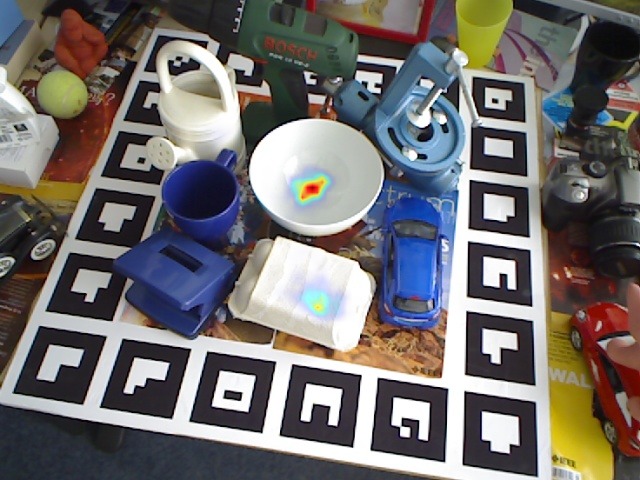}
	\includegraphics[width=3.7cm]{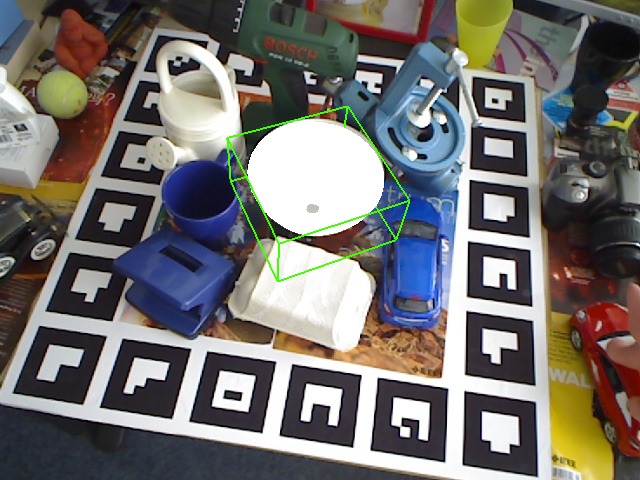}
	\includegraphics[width=3.7cm]{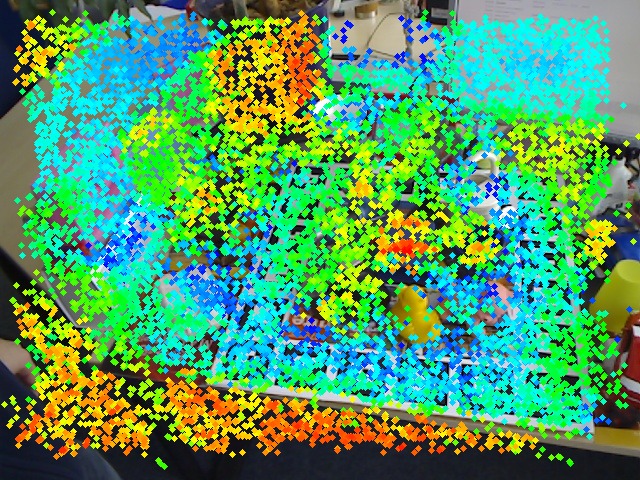}
	\includegraphics[width=3.7cm]{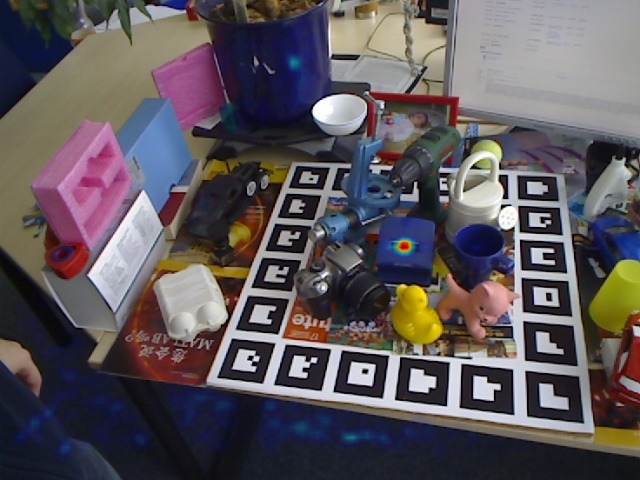}
	\includegraphics[width=3.7cm]{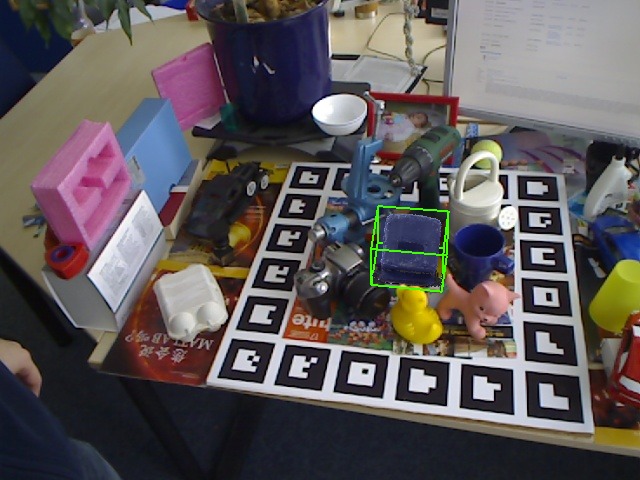}
	\includegraphics[width=3.7cm]{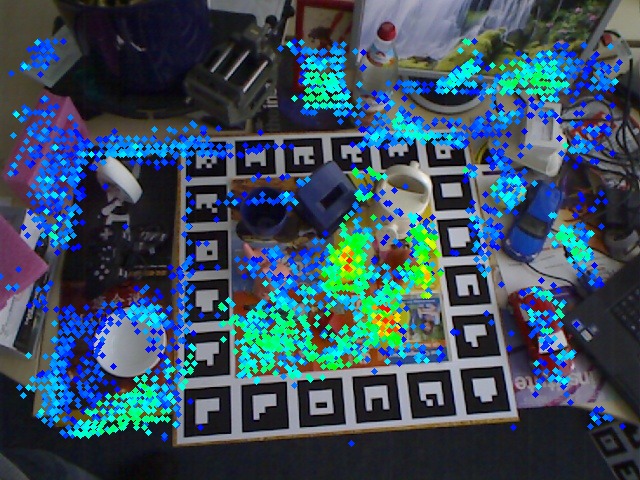}
	\includegraphics[width=3.7cm]{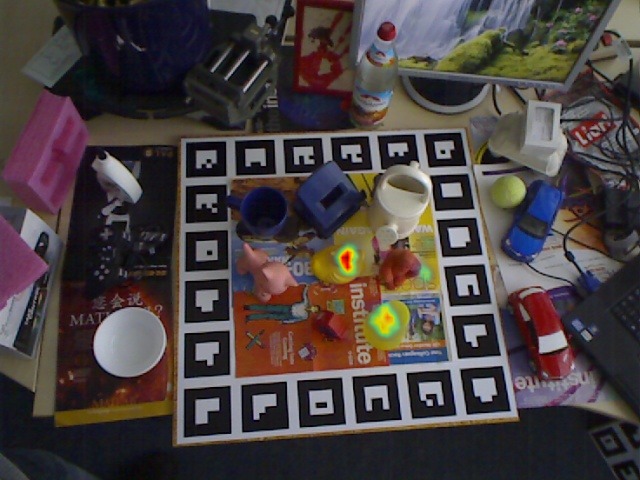}
	\includegraphics[width=3.7cm]{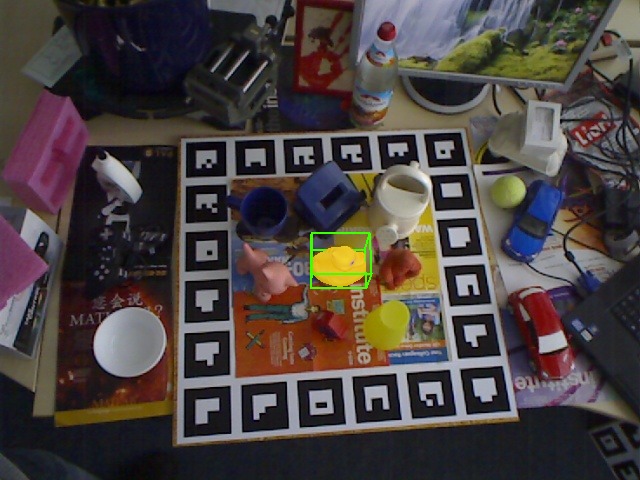}
	\includegraphics[width=3.7cm]{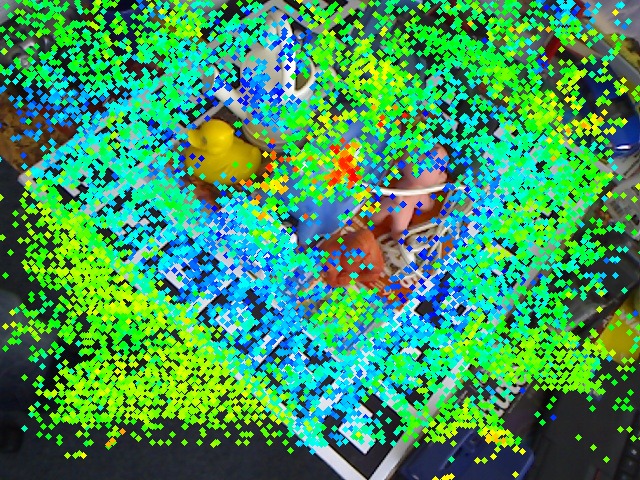}
	\includegraphics[width=3.7cm]{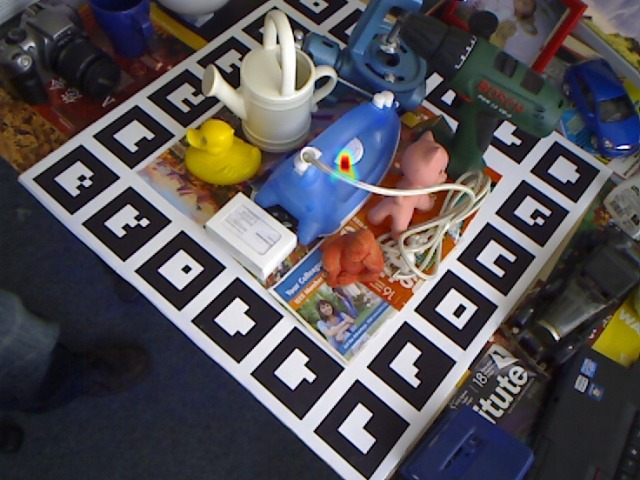}
	\includegraphics[width=3.7cm]{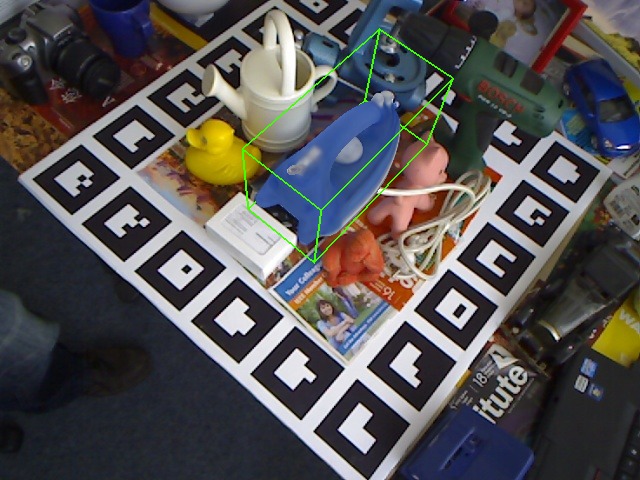}
	\includegraphics[width=3.7cm]{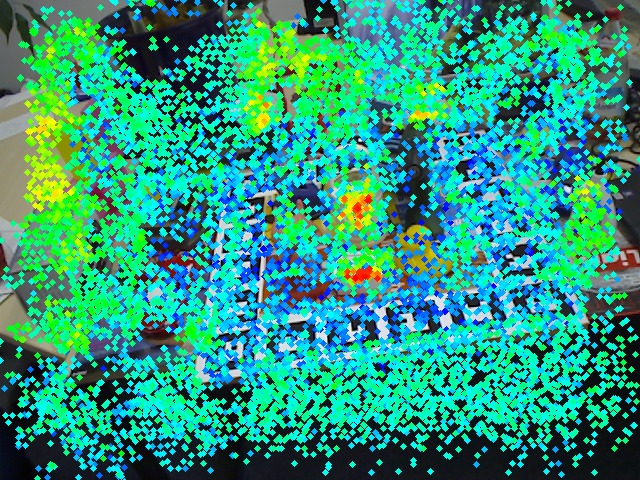}
	\includegraphics[width=3.7cm]{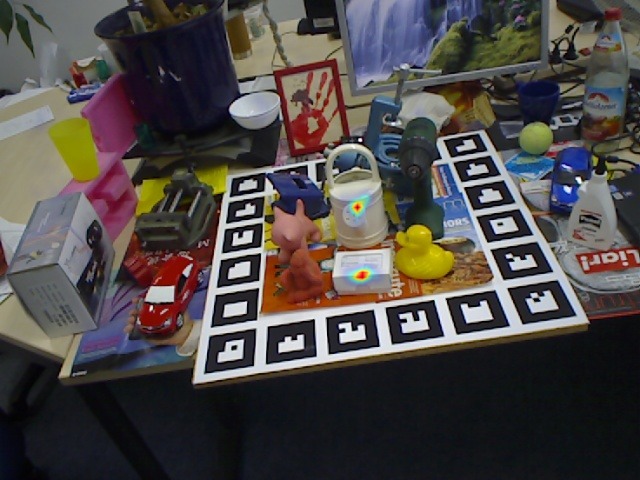}
	\includegraphics[width=3.7cm]{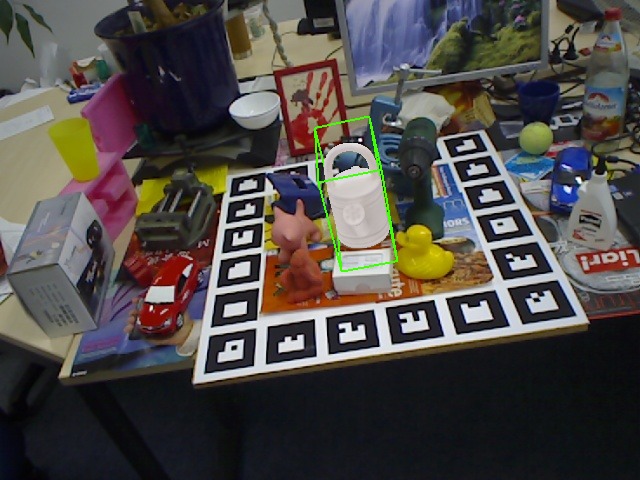}
	\includegraphics[width=3.7cm]{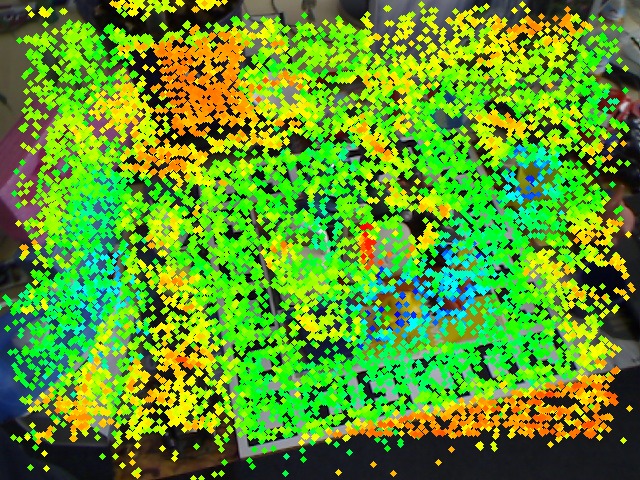}
	\includegraphics[width=3.7cm]{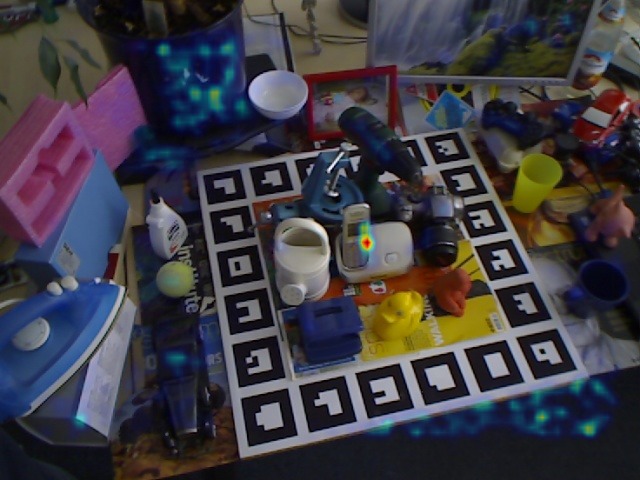}
	\includegraphics[width=3.7cm]{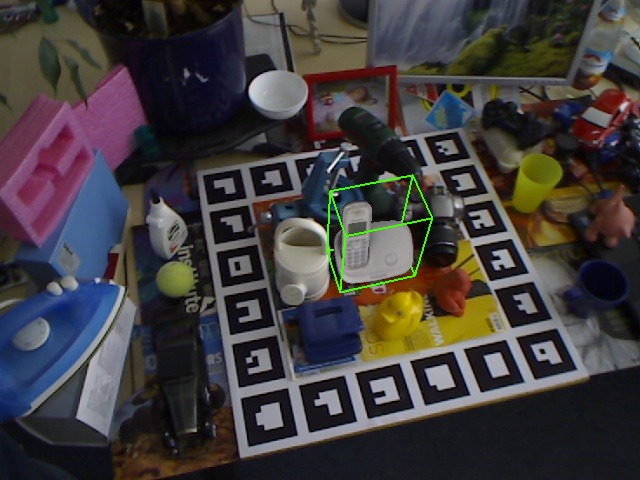}
	\caption{Showing vote maps, probability maps after filtering and detection output on some frames for different objects on the LineMOD dataset.}
	\label{fig:linemod_img}
\end{figure}

We compute the precision average over the 13 objects also used in \cite{Brachmann2014} and report $95.2\%$. We are thus between their plane-trained model with an average of $98.3\%$ and their noise-trained model of $92.6\%$ on pure synthetic data.
We fare relatively well with our detections and can position ourselves nicely between the other state-of-the-art approaches. 
We could observe that we have a near-perfect recall for each object and that our network regresses reliable features allowing to match between synthetic and real local patches. We regard this to be the most important finding of our work since achieving high precision on a dataset can be usually fine-tuned. Nonetheless, the recall for the 'glue' is rather low since it is thin and thus occasionally missed by our sampling.
Based on the overall observation, our comparison of the F1-scores with \cite{Tejani2014} gives further proof of the soundness of our method. We can present excellent numbers and also show some qualitative results of the votes and detections in Figure \ref{fig:linemod_img}.

\begin{table}
	\footnotesize
	\begin{center}
		\begin{tabular}{@{}c|c|c|c|c|c|c|c|c|c|c|c|c|c|c|c@{}}
			& ape 	& bvise  & bowl & cam & can & cat & cup & driller & duck & eggb & glue & holep & iron & lamp & phone \\ \hline
			Us  &  \textbf{96.9}	& 94.1  & \textbf{99.9} &  97.7& 95.2& 97.4 & \textbf{99.6} &  96.2    &  \textbf{97.3}  & 99.9 & 78.6  & 96.8 & \textbf{98.7} & 96.2    & 92.8 \\
			\cite{Hinterstoisser2012} & 95.8 		& 98.7  & \textbf{99.9} & 97.5 & 95.4 & \textbf{99.3} & 97.1 & 93.6 & 95.9 & 99.8 & 91.8 & 95.9 & 97.5 & 97.7 & 93.3 \\
			\cite{Kehl2015}  & 96.1 		& 92.8  & 99.3 & 97.8 & 92.8 & 98.9 & 96.2 & \textbf{98.2} & 94.1 & 99.9 & 96.8 & 95.7 & 96.5 & 98.4 & 93.3 \\
			\cite{Rios-Cabrera2013}  & 95.0 		& 98.9  & 99.7 & \textbf{98.2} & \textbf{96.3} & 99.1 & 97.5 & 94.3 & 94.2 & 99.8 & 96.3 & \textbf{97.5} & 98.4 & 97.9 & \textbf{95.3} \\
			\cite{Hodan2015}  & 93.9		& \textbf{99.8}  & 98.8 & 95.5 & 95.9 & 98.2 & 99.5 & 94.1 & 94.3 & \textbf{100} & \textbf{98.0} & 88.0 & 97.0 & 88.8 & 89.4 \\
		\end{tabular}
	\end{center}
	\caption{Detection rate for each sequence of \cite{Hinterstoisser2012} using the original protocol.}
	\label{table:linemod_prec}
\end{table}

\vspace{-1cm}
\begin{table}
	\small
	\begin{center}
		\begin{tabular}{@{}c|c|c|c|c|c|c|c|c|c|c|c|c|c|c|c@{}}
			& ape & bvise  & bowl & cam & can & cat & cup & driller & duck & eggb & glue & holep & iron & lamp & phone \\ \hline
			Us    & \textbf{98.1}& 94.8   & 100 &  \textbf{93.4} &\textbf{82.6} & \textbf{98.1} & 99.9&  \textbf{96.5} & \textbf{97.9} & \textbf{100}  & \textbf{74.1} & \textbf{97.9}  & \textbf{91.0} &  \textbf{98.2} & \textbf{84.9}     \\
			\cite{Hinterstoisser2012} & 53.3 & 84.6  & - & 64.0 & 51.2 & 65.6 & - & 69.1 & 58.0 & 86.0 & 43.8 & 51.6 & 68.3 & 67.5 & 56.3 \\
			\cite{Tejani2014} & 85.5 		& \textbf{96.1}  & - & 71.8 & 70.9 & 88.8 & - & 90.5 & 90.7 & 74.0 & 67.8 & 87.5 & 73.5 & 92.1 & 72.8 \\
		\end{tabular}
	\end{center}
	\caption{F1-scores for each sequence of \cite{Hinterstoisser2012}. Note that these LineMOD scores are supplied from Tejani et al. with their evaluation since \cite{Hinterstoisser2012} does not provide them. It is evident that our method performs by far better than the two competitors.}
	\label{table:linemod_f1}
\end{table}

\subsection{Challenge dataset}
Lastly, we also evaluated on the 'Challenge' dataset used in \cite{Aldoma2015} containing 35 objects in 39 tabletop sequences with varying amounts of occlusion. The related work usually combines many different cues and descriptors together with elaborate verification schemes to achieve their results. We use this dataset to convey three aspects: we can reliably detect multiple objects undergoing many levels of occlusion while attaining acceptable detection results, we show again generalization on unseen data and that we accomplish this at low runtimes. We present a comparison of our method and related methods in Table 6 together with the average runtime per frame in Figure 11. Since we do not employ a computationally heavy verification the precision of our method is the lowest due to false positives surviving the checks. Nonetheless, we have a surprisingly high recall with our feature regression and voting scheme that brings our F1-score into a favorable position.
It is important to note here that the related works employ a combination of local and global shape descriptors often directly processing the 3D point cloud, exploiting different color, texture and geometrical cues and this taking up to 10-20 seconds per frame. Instead, although our method does not attain such accuracy, it still provides higher efficiency thanks to the use of RGB-D patches only, as well as good scalability with the number of objects due to our discrete sampling (leading to an upper bound on the number of retrieved candidates) and approximate nearest-neighbor retrieval relying on sub-linear methods.

\begin{figure}[t]

	\centering
	\includegraphics[width=2.3cm]{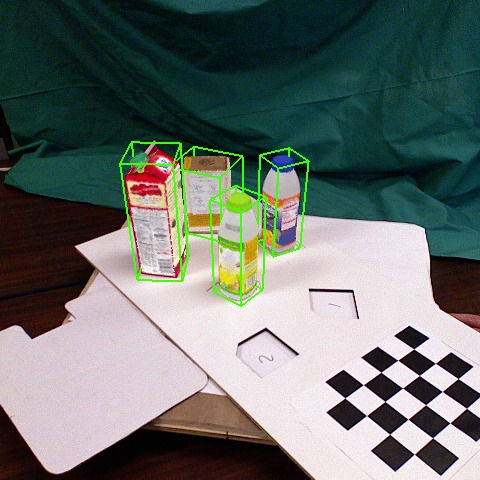}
	\includegraphics[width=2.3cm]{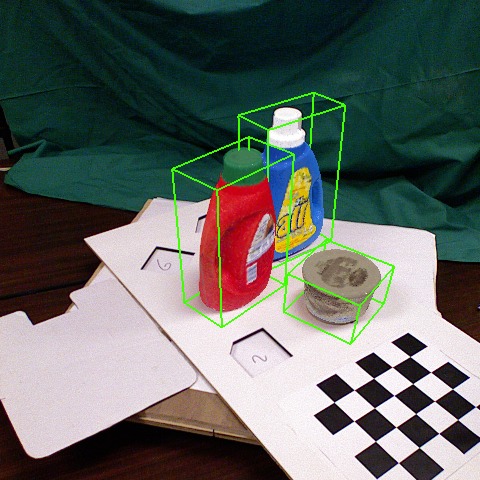}
	\includegraphics[width=2.3cm]{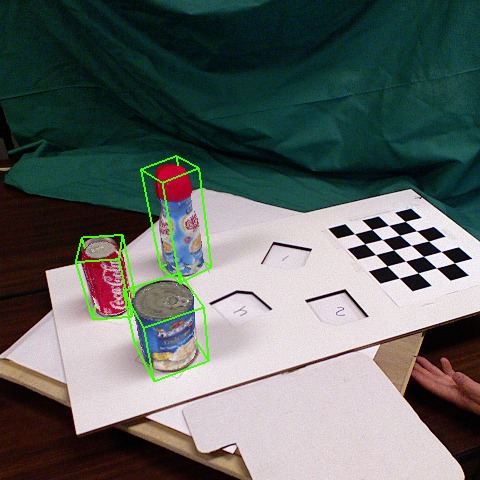}
	\includegraphics[width=2.3cm]{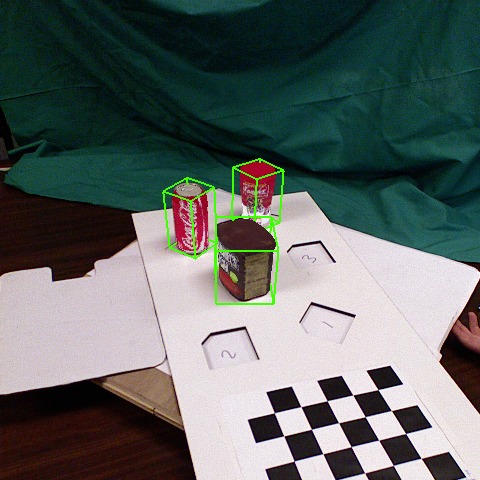}
	\includegraphics[width=2.3cm]{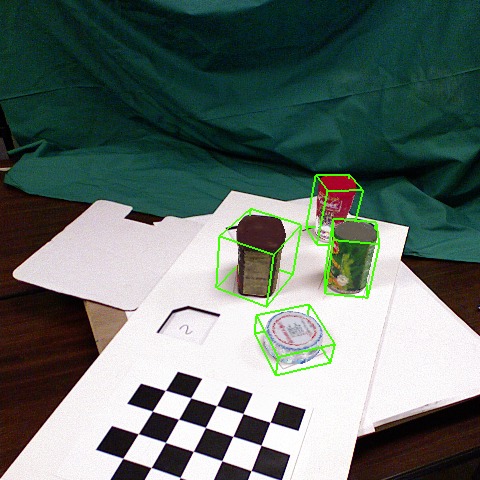}
	\caption{Detection output on selected frames from the 'Challenge' dataset.}
	\label{fig:challenge_img}
\end{figure}

\begin{figure}[t]

	\begin{floatrow}
		\capbtabbox{			
			\begin{tabular}{c|c|c|c}
				Method	& Precision & Recall	& F1-score \\
				\hline	    
				GHV	\cite{Aldoma2015}	& 1.00  & 0.998	&  0.999 \\
				Tang \cite{Tang2011}	& 0.987 & 0.902	&  0.943 \\
				Xie	 \cite{Xie2013}	& 1.00  & 0.998 &  0.999 \\
				Aldoma \cite{Aldoma2013}	& 0.998 & 0.998 &  0.997 \\
				Our approach & 0.941 & 0.973	&  0.956 \\
			\end{tabular}
			\label{fig:challenge_nums} \vspace{1cm}
		}{\caption{Precision, recall and F1-scores on the 'Challenge' dataset. }}
		
		\ffigbox{\includegraphics[width=6cm, height=3.5cm]{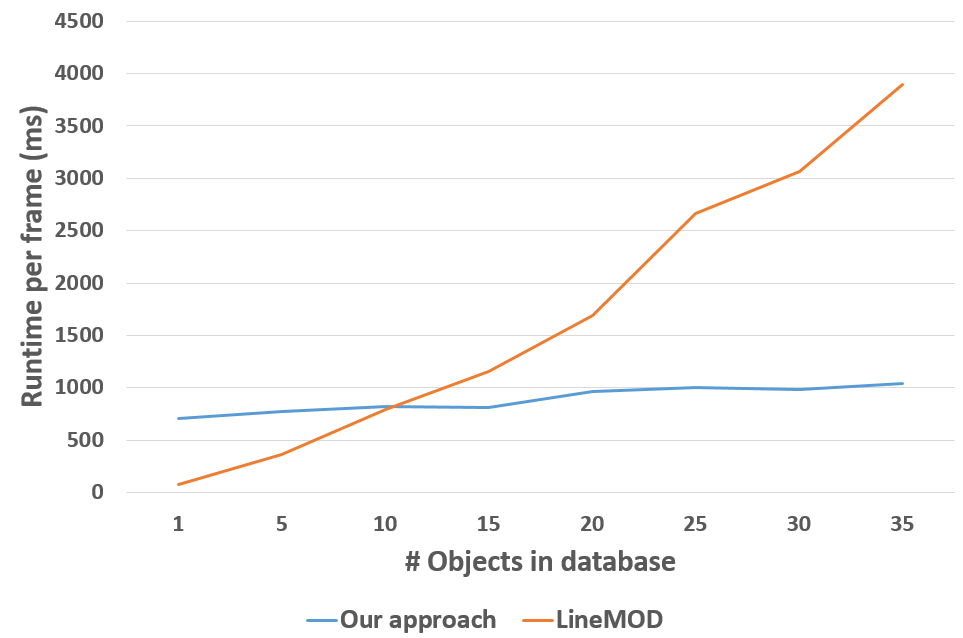}}
		{	\label{fig:challenge_times}	\caption{Average runtime per frame on the 'Challenge' dataset with a changing amount of objects in the database.}}
		
	\end{floatrow}
\end{figure}

\section{Conclusion}
We showed that convolutional auto-encoders have the ability to regress meaningful and discriminative features from local RGB-D patches even for previously unseen input data, facilitating our method and allowing for robust multi-object and multi-instance detection under various levels of occlusion. Furthermore, our vote casting is inherently scalable and the introduced filtering stage allows to suppress many spurious votes. One main observation is that CAEs can abstract enough to reliably match between real and synthetic data. It is still unclear how a more refined training can further increase the results since different architectures have a tremendous impact on the network's performance. 

Another problematic aspect is the complexity of hypothesis verification which can increase exponentially with the amount of objects and votes. Having a method that can combine local and holistic voting together with a learned verification into a single framework could lead to a higher detection precision. A proper in-depth analysis is promising and demands future work. 
\newline 
\textbf{Acknowledgments }
The authors would like to thank Toyota Motor Corporation for supporting and funding this work.

\clearpage

\bibliographystyle{splncs03}
\bibliography{0514}

\clearpage

\chapter*{Supplementary Material}

\section{Self-evaluation with changing parameters}
Our method is mainly governed by three parameters: $\tau$ for constrained voting, $k$ as the number of retrieved neighbors from the codebook, and the sampling density. We ran multiple experiments on the dataset of Tejani et al. and give further insight.

\begin{figure}
	\centering
	\includegraphics[height=3cm]{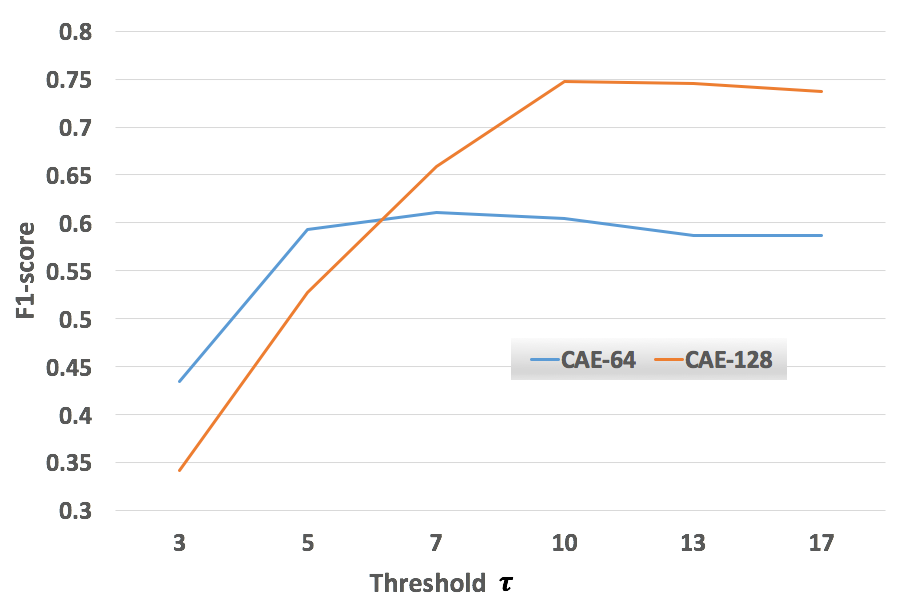}
	\hspace{0.2cm}
	\includegraphics[height=3cm]{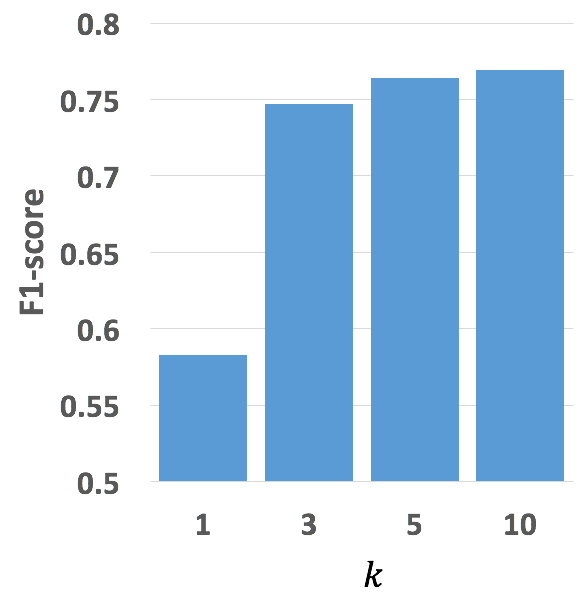}
	\hspace{0.2cm}
	\includegraphics[height=3cm]{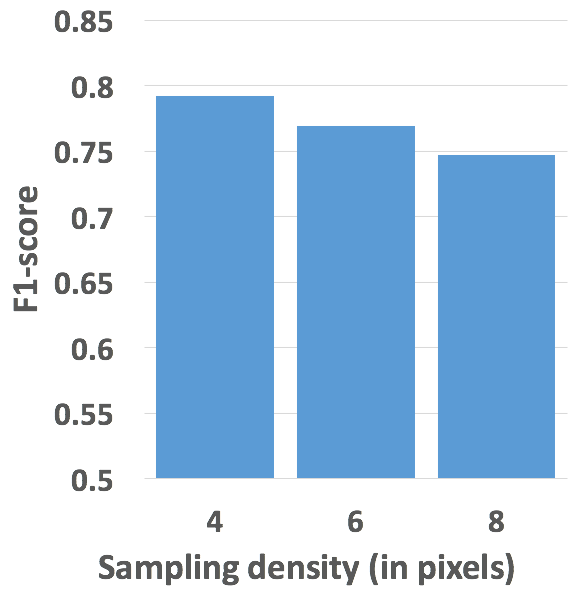}
	\caption{Evaluation of parameter influence. From left to right: threshold $\tau$, number of retrieved neighbors $k$, sampling step size in pixels.}
	\label{fig:self}
\end{figure}

We first wanted to convey the importance of constrained voting via the left graph in Figure \ref{fig:self}. 
Apparently, the threshold needs to reflect the dimensionality of the features, i.e. if the feature is of higher dimensionality, the norm difference $||f(x)-f(y) ||$ grows accordingly. Nonetheless, larger features are more descriptive: while initially both networks underperform since many correct votes are disallowed from being casted, CAE-64 reaches its peak performance already at around $\tau=7$ and from there on additional votes add to more confusion and false positives in the scene. CAE-128 peaks at around $\tau=10$ and shows a similar behavior as CAE-64 for larger thresholds, albeit of smaller effect.

The number of retrieved neighbors $k$ and the change in the F1-score can be seen in the center plot from Figure \ref{fig:self}. Interestingly, the choice of $k$ does not impact our general accuracy too much, apart from the inital jump from $k=1$ to $k=3$. This means that a good match beween real and synthetic data is most often found among the first retrieved neighbors. Furthermore, our verification usually always decides correctly for the geometrically best fitting candidate if multiple hypotheses coincide at the same spatial position (centroids are closer than 5 cm). We also show in the right plot that a denser sampling improves the overall accuracy as expected. This was especially observable for the small "camera" as well as the "shampoo" that exhibits only its thin side at times and can be missed  during sampling.

Unfortunately, with a higher $k$ the runtime increases drastically since the number of hypotheses after mean shift can range in the hundreds per extracted mode. This is due to the fact that we cluster together all votes from the immediate neighbors for each local maximum. In turn, this can lead to multiple seconds of pose refinement and subsequent verification. The same happens with a finer sampling of the scene since the total number of scene votes has an upper bound of $\# samples \cdot k$, extremely cut down by $\tau$ in practice. We therefore fixed $k=3$ and a sampling step of 8 as a reasonable compromise.
\vspace{-0.25cm}
\section{Feature retrieval quality}
\begin{figure}
	\vspace{-0.5cm}
	\centering
	\includegraphics[width=1.35cm]{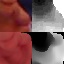}
	\includegraphics[width=1.35cm]{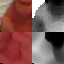}
	\includegraphics[width=1.35cm]{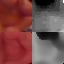}
	\includegraphics[width=1.35cm]{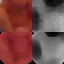}
	\hspace{0.5cm}
	\includegraphics[width=1.35cm]{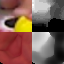}
	\includegraphics[width=1.35cm]{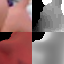}
	\includegraphics[width=1.35cm]{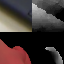}
	\includegraphics[width=1.35cm]{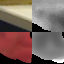}	
	\newline 	\vspace{0.2cm}
	\includegraphics[width=1.35cm]{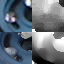}
	\includegraphics[width=1.35cm]{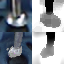}
	\includegraphics[width=1.35cm]{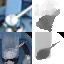}
	\includegraphics[width=1.35cm]{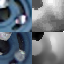}
	\hspace{0.5cm}
	\includegraphics[width=1.35cm]{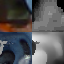}
	\includegraphics[width=1.35cm]{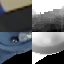}
	\includegraphics[width=1.35cm]{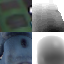}
	\includegraphics[width=1.35cm]{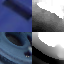}
	\newline 	\vspace{0.2cm}
	\includegraphics[width=1.35cm]{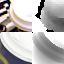}
	\includegraphics[width=1.35cm]{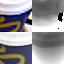}
	\includegraphics[width=1.35cm]{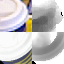}
	\includegraphics[width=1.35cm]{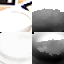}
	\hspace{0.5cm}
	\includegraphics[width=1.35cm]{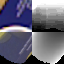}
	\includegraphics[width=1.35cm]{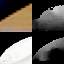}
	\includegraphics[width=1.35cm]{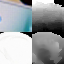}
	\includegraphics[width=1.35cm]{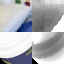}
	\newline 	\vspace{0.2cm}
	\includegraphics[width=1.35cm]{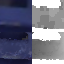}
	\includegraphics[width=1.35cm]{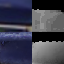}
	\includegraphics[width=1.35cm]{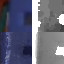}
	\includegraphics[width=1.35cm]{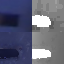}
	\hspace{0.5cm}
	\includegraphics[width=1.35cm]{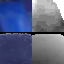}
	\includegraphics[width=1.35cm]{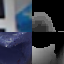}
	\includegraphics[width=1.35cm]{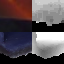}
	\includegraphics[width=1.35cm]{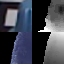}
	\newline 	\vspace{0.2cm}
	\includegraphics[width=1.35cm]{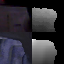}
	\includegraphics[width=1.35cm]{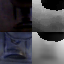}
	\includegraphics[width=1.35cm]{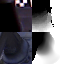}
	\includegraphics[width=1.35cm]{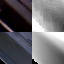}
	\hspace{0.5cm}
	\includegraphics[width=1.35cm]{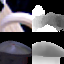}
	\includegraphics[width=1.35cm]{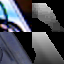}
	\includegraphics[width=1.35cm]{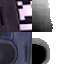}
	\includegraphics[width=1.35cm]{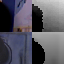}
	\newline 	\vspace{0.2cm}
	\includegraphics[width=1.35cm]{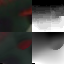}
	\includegraphics[width=1.35cm]{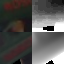}
	\includegraphics[width=1.35cm]{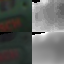}
	\includegraphics[width=1.35cm]{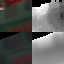}
	\hspace{0.5cm}
	\includegraphics[width=1.35cm]{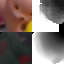}
	\includegraphics[width=1.35cm]{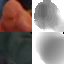}
	\includegraphics[width=1.35cm]{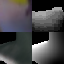}
	\includegraphics[width=1.35cm]{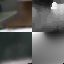}
	\caption{Putative RGB-D patch matches. For each scene input patch, we show the retrieved nearest neighbor from the synthetic model database. For easier distinction, we divided the matches up into correct (left column) and wrong (right column). As can be seen, the features do reflect the visual similarity quite well, even for wrong matches.}
	\label{fig:feats}
\end{figure}

\vspace{-0.25cm}
For a visual feedback of the feature quality we refer to Figure \ref{fig:feats}. For each depicted object we took the first frame of the respective sequence and show the closest neighbor from the codebook ($\tau=\infty$). It is obvious that the features represent well the underlying visual appearance since the putative matches resemble each other well in color and depth.

\end{document}